\documentclass[letterpaper, 10 pt, conference]{ieeeconf}  

\IEEEoverridecommandlockouts    

\usepackage{amssymb} 
\usepackage{makecell} 
\usepackage{multirow}
\usepackage{graphicx}
\usepackage{xspace}    
\let\labelindent\relax
\usepackage{booktabs,enumitem}
\usepackage{amsmath}
\usepackage{caption}
\usepackage{tcolorbox} 

\newcommand{\ours}{\textit{CLON}\xspace}

\title{\LARGE \bf
CLON: Cue-Calibrated Linguistic Object Onboarding for \\Zero-Shot 6D Pose Front-Ends
}

\author{Seojin Ji, Yoojin Kwon and Hyung-Sin Kim$^{1}$
\thanks{$^{1}$Seojin Ji, Yoojin Kwon, and Hyung-Sin Kim are with Seoul National University. Email: \texttt{seojinji23@snu.ac.kr}}
}

\begin{document}

\maketitle
\thispagestyle{empty}
\pagestyle{empty}

\begin{abstract} 
Novel-object 6D pose pipelines depend on front-end hypotheses that retain target instances under clutter and occlusion without overwhelming downstream pose estimation. 
We introduce Cue-Calibrated Linguistic Object Onboarding (\ours), a training-free proposal-and-matching front-end. \ours converts multi-view CAD renderings into compact linguistic memories that provide object-specific conditioning to a frozen proposal generator. Constructed offline and reused across scenes, these memories augment generic prompts before segmentation; subsequent template verification assigns object identity. 
As a complementary onboarding step, \ours derives fixed cue weights from template statistics for semantic, appearance, and geometric verification. 
On seven BOP-Classic-Core datasets, \ours improves detection AP by 8.2 percentage points (pp), segmentation AP by 6.2 pp, and downstream pose AR by up to 4.1 pp over the evaluated CNOS and SAM-6D front-ends. 
In a real-world localization-to-grasp study under clutter and occlusion with 100 trials per method, \ours's improved localization doubles grasp success relative to the strongest evaluated baseline. Controlled ablations evaluate linguistic conditioning, multi-view construction, and verification. Analysis of consolidated SAM~3 proposals before verification further shows a smaller candidate set with higher class-agnostic recall on five datasets.

\end{abstract}


\section{Introduction}\label{sec:intro}

Vision-based 6D object pose estimation is a core capability for robots that grasp, manipulate, inspect, or assemble objects. 
Recent render-and-compare and foundation model-based methods have made substantial progress on \textit{novel} object pose estimation by requiring only CAD models~\cite{labbe2022megapose, lin2024sam,nguyen2024gigapose} or reference views~\cite{liu2022gen6d, wen2024foundationpose, lee2025any6d} without object-specific training. 
In many modular pipelines, however, the pose solver receives only \textit{hypotheses} supplied by a \textbf{proposal-and-matching front-end}~\cite{lin2024sam, nguyen2023cnos}. Missing a target leaves no valid hypothesis for pose recovery, while excessive candidates increase downstream computation. Front-end quality therefore remains important even as pose solvers improve~\cite{nguyen2025bop}.

Many render-and-match front-ends build a bank of multi-view CAD renderings and use it to identify candidate regions after generic proposals have been generated~\cite{lin2024sam,nguyen2023cnos}. In these pipelines, the available target-specific information does not condition proposal generation. 
Concept-conditioned segmenters such as SAM 3~\cite{carion2025sam}  provide an opportunity to change this: linguistic descriptions can influence which regions are proposed. 
However, generic prompts such as ``object'' and ``small object'' do not express the properties of the onboarded objects. The challenge is to convert multi-view CAD evidence during onboarding into \textit{reusable conditioning} for online proposal generation, rather than simply replace the segmenter or improve subsequent matching.

%

We introduce Cue-Calibrated Linguistic Object Onboarding (\ours), a training-free proposal-and-matching front-end. 
Its main contribution is to make onboarding configure proposal generation, rather than only prepare templates for subsequent matching. \ours converts multi-view CAD renderings into \textbf{compact object-specific linguistic memories} that provide reusable conditioning for the proposal generator.
Constructed offline and reused across scenes, these linguistic memories augment generic prompts during proposal generation without manually written object descriptions or model updates. Each retained proposal is then compared with all onboarded CAD objects, independently of the prompt that generated it. Language therefore guides search, while template verification assigns identity. 
As a complementary onboarding step, \ours computes \textbf{object-set cue weights} from template statistics offline and applies these fixed weights to combine semantic, appearance, and geometric verification scores during online inference.
Across seven BOP datasets, \ours improves detection AP by 8.2 pp, segmentation AP by 6.2 pp, and downstream pose AR by up to 4.1 pp over the evaluated front-ends. In a real-world localization-to-grasp evaluation under clutter and occlusion, \ours further doubles grasp success relative to the strongest evaluated baseline.
With the verifier fixed, linguistic memory raises detection AP by 14.1 pp with SAM 3 and 13.7 pp with Grounding DINO + SAM. Separate controlled comparisons further confirm the benefits of multi-view onboarding and template-derived cue weighting.
%
Before verification, \ours also produces substantially fewer proposals with higher class-agnostic recall on the five datasets with recall measurements.

Our contributions are threefold. 
\begin{itemize}[leftmargin=*]
    \item We introduce linguistic object onboarding that converts multi-view CAD templates into reusable conditioning for proposal generation with frozen pretrained models. 
    \item We derive object-set-specific cue weights from rendered-template statistics during onboarding and keep them fixed throughout online scene inference.
    \item We evaluate \ours on seven BOP-Classic-Core datasets with three downstream pose solvers and a robot localization-to-grasp study, with controlled analyses of linguistic memory, multi-view construction, and cue calibration. 
\end{itemize}

\begin{figure*}[t]
    \centering
    \includegraphics[width=\linewidth]{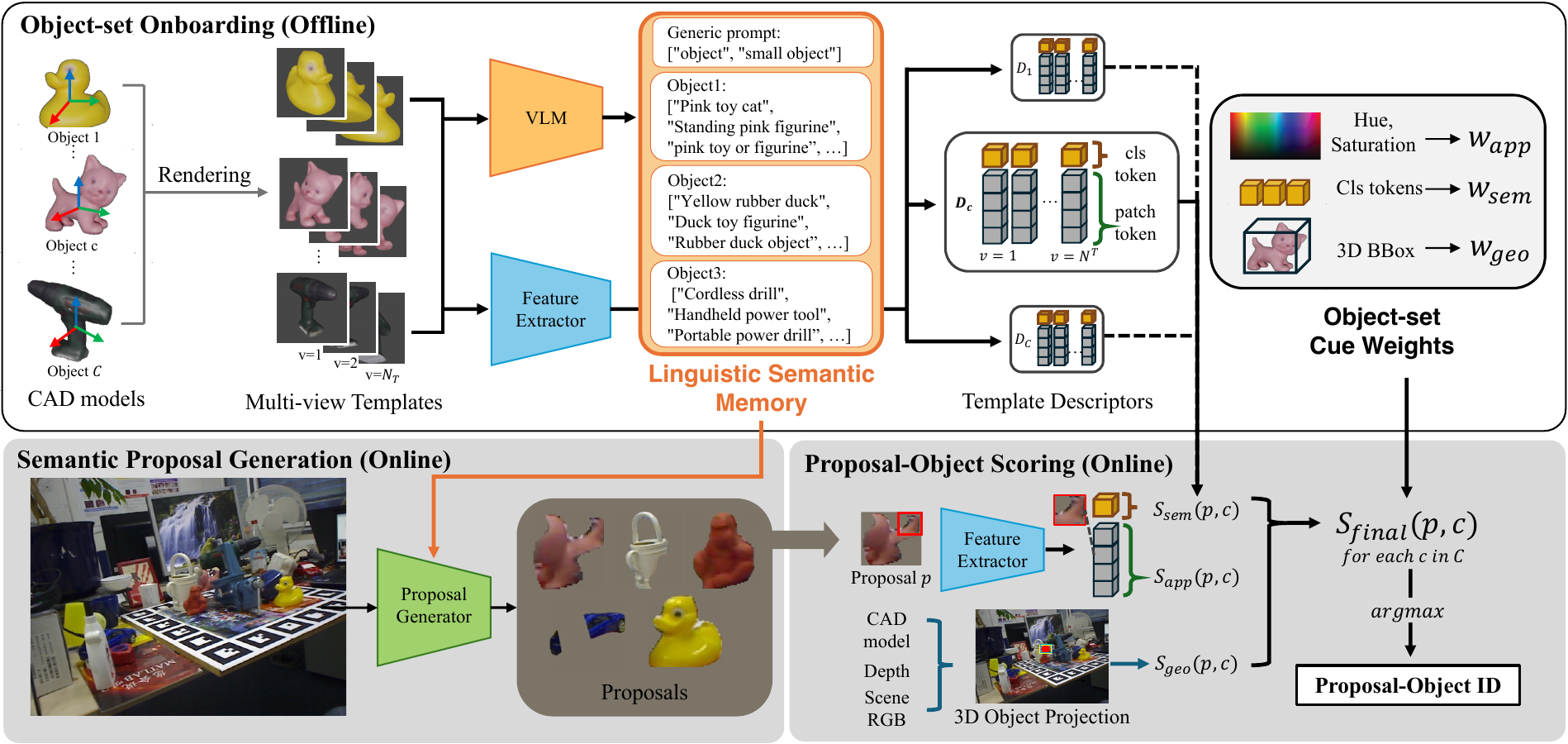}
    \caption{\textbf{Method Overview.}      
    Offline onboarding builds linguistic object memories for target-aware proposal generation and object-set cue weights for calibrated proposal-object scoring.     
    }
    \label{fig:method}
\end{figure*}

\section{Related Work}\label{sec:related work}

\subsection{Promptable Segmentation for Semantic Object Search}

%
The Segment Anything Model (SAM)~\cite{kirillov2023segment} and SAM 2~\cite{ravi2025sam} generate masks from spatial prompts. 
%
GroundedSAM~\cite{ren2024grounded} and OpenWorldSAM~\cite{openworldsam2025} connect segmentation to language-based search, while Semantic-SAM~\cite{li2024segment} supports semantic recognition and multi-granularity segmentation.
SAM 3~\cite{carion2025sam} further introduces concept-conditioned segmentation through noun phrases and visual exemplars. 
These methods provide interfaces for specifying what to segment.
\ours complements them through CAD-to-language onboarding: a frozen VLM~\cite{bai2025qwen25vltechnicalreport} converts rendered object views into \textit{reusable linguistic prompts}. Its intervention is the construction of object-specific conditioning, rather than the segmentation model itself.

\subsection{Template-Based Object Detection and Segmentation}

Novel-object pose estimation pipelines require a front-end that segments, detects and identifies unseen object instances from CAD models or reference views. 
CNOS~\cite{nguyen2023cnos} combines generic SAM proposals with DINOv2-based~\cite{oquab2023dinov2} template matching.
SAM-6D~\cite{lin2024sam} extends this idea by combining semantic, appearance, and geometric evidence, while MUSE~\cite{cho2026muse} improves proposal scoring through relative similarity and an uncertainty-aware object prior.
%
NIDS-Net~\cite{lu2025adapting} combines Grounding DINO~\cite{liu2024grounding} with SAM for proposal generation. Although the generator supports language, NIDS-Net uses the generic query ``objects''; instance-specific information is used mainly during matching. 
\ours instead uses onboarding-derived object descriptions to \textit{condition proposal generation itself}.

%


Object-conditioned prediction is also explored by OC-DiT~\cite{ulmer2025ocdit}, which conditions a latent diffusion model on object templates to generate target instance masks. 3PT~\cite{Kalra20263PT} directly conditions a unified detection, segmentation, and pose framework on 3D object models.
\ours instead constructs reusable linguistic conditioning for \textit{frozen} pretrained components while retaining template-based verification.

\subsection{Novel-Object 6D Pose Estimation}

MegaPose~\cite{labbe2022megapose} estimates pose through render-and-compare refinement, GigaPose~\cite{nguyen2024gigapose} uses template retrieval and local correspondences, and FoundationPose~\cite{wen2024foundationpose} supports model-based and model-free pose estimation and tracking. FoundPose~\cite{ornek2024foundpose} and FreeZe~\cite{caraffa2024freeze} exploit foundation features for training-free pose estimation. 
These methods primarily address pose recovery. \ours studies the \textit{hypotheses} supplied to that stage (i.e., the front-end) and is evaluated with GigaPose, SAM-6D, and FoundationPose while leaving their pose-estimation components unchanged.

%

\section{Method} \label{sec:method}

Fig.~\ref{fig:method} summarizes \ours. Given an onboarded object set $\mathcal C$, Cue-Calibrated Linguistic Object Onboarding (\ours) separates \textit{offline object onboarding} from \textit{online scene inference}. Each object instance $c\in\mathcal C$ is represented by rendered templates $T_c=\{(I_c^v,D_c^v)\}_{v=1}^{N_T}$, where $v$ indexes a template view while $I_c^v$ and $D_c^v$ denote its RGB image and depth map, respectively.
During onboarding, \ours constructs an object-set memory
$M_{\mathcal C}=\big(\{L_c\}_{c\in\mathcal C},w_{\mathcal C}\big)$. 
Each linguistic object memory $L_c$ is generated from rendered object views and optionally augmented with the coarse metadata described below, while object-set-specific cue weights $w_{\mathcal C}=(w_{\rm sem},w_{\rm app},w_{\rm geo})$ are computed from template statistics. Both are fixed before scene inference.
%
During online inference, \ours uses $L_c$ to guide proposal generation toward target-relevant regions and $w_{\mathcal C}$ to calibrate semantic, appearance, and geometric cues for proposal--object scoring. The resulting hypotheses are then passed to the downstream pose estimator.


\subsection{Object-Set Onboarding}
\label{sec:onboarding}

\noindent\textbf{Linguistic Semantic Memory.} \quad
The primary role of \ours's offline onboarding is to convert multi-view CAD
evidence into compact object-specific linguistic queries that can guide
proposal generation before template matching. For each object $c$, we
subsample $N'_T$ template views ($N'_T < N_T$) from $T_c$ at regular
angular intervals to capture complementary visual characteristics across
viewpoints. Then we query Qwen-2.5-VL-7B-Instruct~\cite{bai2025qwen25vltechnicalreport} using a \textit{fixed textual prompt} in Fig.~\ref{fig:prompt_config}. 
The prompt contains an optional high-level metadata field, such as ``industrial object'' or ``household object,'' when such metadata is available during onboarding; otherwise the field is left empty. The same prompting rule is used before scene inference and is never adjusted using test scenes. 
%

The prompt asks the VLM to generate $M$ short noun phrases that capture complementary and persistent object properties. Rather than relying on a
single object name, the prompt encourages specific identity terms, spatial or boundary related descriptors when informative, distinctive object parts, and color cues when useful. This design provides multiple linguistic descriptions that can activate different visual evidence for the same target under viewpoint change, partial visibility, or clutter.

We use the parsed phrases directly as the linguistic semantic memory $L_c$,
without human filtering. During online inference, the phrases in $L_c$ serve
as object-specific proposal queries together with the generic prompt set
$T_{\rm gen}$. Importantly, proposals are not assigned the identity of the
object whose phrase generated them. After proposal generation and
consolidation, every retained proposal is verified against all onboarded
objects. Thus, linguistic memory guides where the front-end searches for
target-relevant regions, while template-based verification determines final
object identity.

%

\begin{figure}[h]
    \centering
    \begin{tcolorbox}[
        title=Prompt Configuration,
        colback=white,
        colframe=black,
        sharp corners,
        before skip=1pt,
        after skip=1pt,
        top=2pt,
        bottom=2pt,
        left=4pt,
        right=4pt,
        boxsep=2pt
    ]
    \small
    \textbf{[Task]}\\
    Identify the 3D object in the rendered views and generate short linguistic
    candidates describing its identity and physical extent.
    
    \vspace{0.4ex}
    \textbf{[Rules]}\\
    1) List exactly four identity candidates.\\
    2) Use the most specific everyday name; avoid vague terms.\\
    3) Include spatial descriptors (e.g.,``Rear,'' ``Hidden,'' or ``Tucked'') when informative.\\
    4) Use boundary/part descriptors that clarify physical extent.\\
    5) Color may be included when useful but is not required in every phrase.\\
    6) Each candidate must contain at most four words.\\
    7) The final candidate must be ``[Dominant Color] object.''\\
    8) Output only the candidate list.
    
    \vspace{0.4ex}
    \textbf{[Example]}: Rear wooden bookshelf; Entire brown storage-unit; Vertical side-panel; Brown object
    \end{tcolorbox}
    \caption{Summary of the fixed prompt used to generate linguistic object memories with Qwen-2.5-VL~\cite{bai2025qwen25vltechnicalreport}.}
    \label{fig:prompt_config}
\end{figure}

\noindent\textbf{Object-Set Cue Weights.} \quad
The verifier in Sec.~\ref{sec:inference} combines semantic, appearance, and geometric cues. Instead of using the same cue weights across all datasets, we derive fixed cue weights for each onboarded object set from statistics of its rendered templates.

Let $P_c^v(h,s)$ denote the empirical joint distribution over Hue ($h$) and Saturation ($s$) bins of the masked template image $I_c^v$ after conversion to HSV. We define the object-set template chromatic entropy 
\begin{equation}
\begin{aligned}
e_{\rm app}
&=
\frac{1}{|\mathcal C|N_T}
\sum_{c\in\mathcal C}
\sum_{v=1}^{N_T}
H(I_c^v), \\
\text{where }H(I_c^v)
&=
-\frac{1}{\log_2 N_B}
\sum_h\sum_s
P_c^v(h,s)
\log_2 P_c^v(h,s).
\end{aligned}
\end{equation}
Here $H(I^v_c)$ is the normalized template entropy, $N_B$ is the number of bins in the joint $(h,s)$ histogram, and zero-probability terms are omitted. 

We use template chromatic entropy to choose between semantic-only verification and a mixed-cue prior. If $e_{\rm app}>\tau_{\rm app}$, the verifier uses semantic-only scoring: $(w_{\rm sem},w_{\rm app},w_{\rm geo})=(1,0,0)$. 
Otherwise, \ours appearance receives weight  \(e_{\mathrm{app}}\), and $\alpha$ divides the remaining weight between semantic and geometric evidence: 
$$(w_{\rm sem}, w_{\rm app}, w_{\rm geo}) = (\alpha (1 - e_{\rm app}),\; e_{\rm app},\; (1 - \alpha)(1 - e_{\rm app}))$$ 
%
As $e_{\rm app}$ decreases, the semantic and geometric cues receive a larger share of the remaining weight.

The semantic--geometry split $\alpha$ is computed from template-derived inter-object discriminability. 
Semantic discriminability $s_{\rm dis}$ is 
$$\begin{aligned}
s_{\rm dis} &= 1 - \frac{1}{|\mathcal{C}|} \sum_{c=1}^{\mathcal{C}} \max_{k \neq c} \left( \frac{g'_c \cdot g'_k}{\|g'_c\| \|g'_k\|} \right), \\
&\text{where } g'_c = \frac{1}{N_T} \sum_{v=1}^{N_T} g^v_c.
\end{aligned}$$
Here $g'_c$ is the mean DINOv3 class-token prototype of the object $c$ across its template views. Hence, a larger $s_{\rm dis}$ indicates greater semantic separation from the nearest competing object.
Geometric discriminability $g_{\rm dis}$ is computed from 3D bounding-box overlap after centering all boxes in their canonical object frames and converting them to a common unit:
$$g_{\rm dis} = 1 - \frac{1}{|\mathcal{C}|} \sum_{c=1}^{\mathcal{C}}\max_{k \neq c} \text{IoU}_{3D}(B_c, B_k)$$
Similarly, a larger $g_{\rm dis}$ indicates greater differences in 3D object dimensions among the onboarded objects.
We set
\[
\alpha
=
\frac{\lambda_{\rm sem}s_{\rm dis}}
{\lambda_{\rm sem}s_{\rm dis}+g_{\rm dis}+\epsilon},
\]
where $\lambda_{\rm sem}$ is a fixed global semantic-priority constant and
$\epsilon$ prevents division by zero. 
For a single-object onboarding set, where inter-object discriminability is undefined, we use semantic-only scoring with $(w_{\rm sem}, w_{\rm app}, w_{\rm geo})=(1,0,0)$.

All cue weights are computed from the onboarded templates only and are kept fixed during online inference, serving as \textbf{proposal--object scoring weights} for calibrating semantic, appearance and geometric evidence.

\subsection{Online Scene Inference}
\label{sec:inference}

\noindent\textbf{Semantic Proposal Generation.} \quad 
Given an observed RGB scene image $I$, \ours combines object-specific linguistic queries with the generic queries ``object'' and ``small object,'' 
so that proposal generation is not restricted to the concepts expressed by the generated descriptions.
We define the prompt set as
$Q=T_{\rm gen}\cup\bigcup_{c\in\mathcal{C}}L_c$ 
where $T_{\rm gen}$=\{``object'', ``small object''\} is the generic prompt set and $L_c$ is the target-specific linguistic prompt set for each object $c \in \mathcal{C}$. 
Each phrase in $Q$ serves as a proposal query to SAM~3. Queries are batched for efficient execution, while the SAM~3 image features are computed once per scene and shared across all generic and object-specific queries. The resulting masks are pooled and consolidated using IoU-based non-maximum suppression:

$$
\mathcal{P}
=
\operatorname{NMS}_{\tau_{\rm iou}}
\left(
\bigcup_{q\in Q}\operatorname{SAM3}(I,q)
\right).
$$

Each retained proposal $p\in\mathcal P$ is represented by a mask $m_p$ and bounding box $B_p$. Proposals are not assigned object identities at this stage; instead, every retained proposal is scored against all onboarded objects, regardless of the prompt from which it originated. Thus, linguistic object memory acts before matching by guiding SAM~3 toward target-relevant regions, while final object identity is determined by the proposal--object verifier.



\noindent\textbf{Calibrated Proposal-Object Scoring.} \quad
For each proposal $p\in\mathcal P$ and object $c\in\mathcal C$, \ours computes a proposal-object score. We reuse the semantic, appearance, and geometric score definitions from SAM-6D~\cite{lin2024sam}, but apply them in a joint proposal-object scoring policy with onboarding-time cue weights.
The semantic score $S_{\rm sem}(p,c)$ measures global compatibility between proposal $p$ and rendered templates of object $c$. The appearance score $S_{\rm app}(p,c)$ measures visual compatibility between the proposal crop and the object templates. The geometric score $S_{\rm geo}(p,c)$ provides coarse spatial consistency when valid depth is available; otherwise, its weight is set to zero and remaining cue weights are renormalized to sum to one. If no weighted cue remains, we fall back to semantic-only scoring.

The final proposal-object score is 
\begin{equation}
\begin{aligned}
S_{\mathrm{final}}(p,c)
&=
\bar w_{\mathrm{sem}}S_{\mathrm{sem}}(p,c)
+
\bar w_{\mathrm{app}}S_{\mathrm{app}}(p,c) \\
&\quad +
\bar w_{\mathrm{geo}}S_{\mathrm{geo}}(p,c).
\end{aligned}
\end{equation}
where $\bar w_j$ are the onboarding-time cue weights renormalized over the available cues. Each proposal is assigned to the object with the highest final score,
\[
c^\ast(p)=\arg\max_{c\in\mathcal C}S_{\rm final}(p,c).
\]
All matched hypotheses satisfying $S_{\rm final}\bigl(p,c^\ast(p)\bigr)>\tau_{\rm match}$ are forwarded to the downstream 6D pose estimator. \ours does not impose an additional top-$K$ cap after matching; proposal generation guided by linguistic memory, followed by verification against onboarded objects, already sufficiently constrains the final hypothesis set.


\section{Experiments}\label{sec:experiments}

\begin{table*}[!ht]
    \centering
    \caption{2D detection and segmentation results on the seven core BOP datasets. The best and second-best results within each dataset are shown in \textbf{bold} and \underline{underline}, respectively.
    }
    \label{tab:detection_segmentation}
    \resizebox{\textwidth}{!}{%
    \begin{tabular}{lcccccccc}
    \toprule
    \multirow{2}{*}{\textbf{Method}} & LM-O~\cite{brachmann2014learning} & T-LESS~\cite{hodan2017t} & TUD-L~\cite{hodan2018bop} & IC-BIN~\cite{doumanoglou2016recovering} & ITODD~\cite{drost2017introducing} & HB~\cite{kaskman2019homebreweddb} & YCB-V~\cite{xiang2017posecnn} & $AP_{\text{mean}}$ \\
    \cmidrule(lr){2-8} \cmidrule(lr){9-9}
    \cmidrule(lr){2-8} \cmidrule(lr){9-9}
    
    \multicolumn{9}{c}{\rule{0pt}{1ex}\textit{Detection Results}} \\ 
    \midrule
    CNOS (SAM)~\cite{nguyen2023cnos} & 0.392 & 0.329 & 0.376 & 0.204 & 0.314 & 0.424 & 0.495 & 0.362 \\
    CNOS (FastSAM)~\cite{nguyen2023cnos} & 0.430 & 0.397 & 0.540 & 0.225 & 0.322 & 0.518 & 0.573 & 0.429 \\
    SAM-6D (SAM)~\cite{lin2024sam} & \underline{0.465} & 0.437 & 0.536 & \underline{0.261} & 0.393 & 0.530 & 0.517 & 0.448 \\
    SAM-6D (FastSAM)~\cite{lin2024sam} & 0.464 & \underline{0.458} & \underline{0.573} & 0.245 & \underline{0.419} & \underline{0.551} & \underline{0.589} & \underline{0.471} \\
    \cmidrule(lr){1-9}
    \textbf{\ours} & \textbf{0.529} & \textbf{0.552} & \textbf{0.665} & \textbf{0.358} & \textbf{0.465} & \textbf{0.628} & \textbf{0.671} & \textbf{0.553}\\ 
    \midrule
    
    \multicolumn{9}{c}{\rule{0pt}{1ex}\textit{Segmentation Results}} \\ 
    \midrule
    CNOS (SAM)~\cite{nguyen2023cnos} & 0.393 & 0.396 & 0.399 & 0.282 & 0.284 & 0.478 & 0.597 & 0.404 \\
    CNOS (FastSAM)~\cite{nguyen2023cnos} & 0.395 & 0.376 & 0.486 & 0.267 & 0.251 & 0.514 & 0.602 & 0.413 \\
    SAM-6D (SAM)~\cite{lin2024sam} & \underline{0.460} & \underline{0.451} & \underline{0.568} & \underline{0.358} & \underline{0.331} & \underline{0.593} & 0.605 & \underline{0.481} \\
    SAM-6D (FastSAM)~\cite{lin2024sam} & 0.423 & 0.420 & 0.517 & 0.294 & 0.319 & 0.548 & \underline{0.621} & 0.449 \\
    \cmidrule(lr){1-9}
    \textbf{\ours} & \textbf{0.488} & \textbf{0.535} & \textbf{0.619} & \textbf{0.465} & \textbf{0.363} & \textbf{0.640} & \textbf{0.692} & \textbf{0.543}\\ 
    \bottomrule
    \end{tabular}%
    }
\end{table*}

\begin{table*}[t]
    \centering
    \caption{6D pose estimation results using different detection/segmentation and pose estimation methods on the seven core BOP datasets. ($\dagger$ denotes the use of MegaPose refinement~\cite{labbe2022megapose}.) The best and second-best results within each dataset are shown in \textbf{bold} and \underline{underline}, respectively.
    }
    \label{tab:pose_estimation_results}
    \setlength{\tabcolsep}{8pt}
    \resizebox{\textwidth}{!}{%
    \begin{tabular}{llcccccccc}
        \toprule
        \textbf{Method} & \makecell[l]{\textbf{Detection/}\\\textbf{Segmentation}} & LM-O & T-LESS & TUD-L & IC-BIN & ITODD & HB & YCB-V & $AR_{\text{mean}}$ \\
        \midrule
        \multirow{4}{*}{GigaPose$\dagger$~\cite{nguyen2024gigapose}} & CNOS (SAM)~\cite{nguyen2023cnos} & 0.524 & 0.547 & 0.472 & 0.456 & 0.373 & 0.600 & 0.640 & 0.516 \\
        & CNOS (FastSAM)~\cite{nguyen2023cnos} & 0.599 & 0.571 & 0.635 & 0.472 & 0.397 & 0.721 & 0.664 & 0.580 \\
         & SAM-6D (SAM)~\cite{lin2024sam} & \underline{0.607} & \underline{0.578} & \underline{0.651} & \underline{0.521} & \underline{0.409} & \underline{0.728} & 0.656 & \underline{0.593} \\
         & SAM-6D (FastSAM)~\cite{lin2024sam} & 0.600 & 0.576 & 0.647 & 0.477 & 0.407 & 0.721 & \underline{0.667} & 0.585 \\
         \cmidrule(lr){2-10}
         & \textbf{\ours} & \textbf{0.620} & \textbf{0.673} & \textbf{0.685} & \textbf{0.567} & \textbf{0.414} & \textbf{0.778} & \textbf{0.699} & \textbf{0.634} \\ 
         \midrule
        \multirow{4}{*}{SAM-6D~\cite{lin2024sam}} & CNOS (SAM)~\cite{nguyen2023cnos} & 0.607 & 0.475 & 0.650 & 0.498 & 0.505 & 0.619 & 0.788 & 0.592 \\
        & CNOS (FastSAM)~\cite{nguyen2023cnos} & 0.649 & 0.483 & 0.824 & 0.496 & 0.545 & 0.734 & 0.819 & 0.650\\
         & SAM-6D (SAM)~\cite{lin2024sam} & \underline{0.700} & \underline{0.514} & \underline{0.892} & \underline{0.585} & \underline{0.582} & \underline{0.766} & \underline{0.835} & \underline{0.696} \\
         & SAM-6D (FastSAM)~\cite{lin2024sam} & 0.667 & 0.489 & 0.836 & 0.515 & 0.569 & 0.733 & 0.831 & 0.663 \\
         \cmidrule(lr){2-10}
         & \textbf{\ours} & \textbf{0.705} & \textbf{0.583} & \textbf{0.894} & \textbf{0.656} & \textbf{0.583} & \textbf{0.790} & \textbf{0.868} & \textbf{0.726} \\ 
         \midrule
        \multirow{4}{*}{FoundationPose~\cite{wen2024foundationpose}} & CNOS (SAM)~\cite{nguyen2023cnos} & 0.633 & 0.485 & 0.691 & {0.567} & {0.585} & 0.677 & 0.851 & 0.641 \\
        & CNOS (FastSAM)~\cite{nguyen2023cnos} & 0.670 & 0.580  & 0.871 & 0.588 & 0.617 & 0.805 & 0.865 & 0.714 \\
         & SAM-6D (SAM)~\cite{lin2024sam} & \underline{0.727} & 0.485 & \textbf{0.898} & \underline{0.636} & \textbf{0.667} & \textbf{0.825} & \underline{0.867} & \underline{0.729} \\
         & SAM-6D (FastSAM)~\cite{lin2024sam} & 0.703 & \textbf{0.599} & 0.873 & 0.601 & \underline{0.649} & 0.802 & \underline{0.867} & 0.728 \\
         \cmidrule(lr){2-10}
         & \textbf{\ours} & \textbf{0.728} & \underline{0.586} & \underline{0.878} & \textbf{0.691} & 0.644 & \underline{0.811} & \textbf{0.896} & \textbf{0.748} \\ 
         \bottomrule
    \end{tabular}%
}
\end{table*}

\subsection{Implementation Detail}

\noindent\textbf{Datasets \& Metrics.\quad} We evaluate \ours on the seven core BOP challenge datasets (LM-O~\cite{brachmann2014learning}, T-LESS~\cite{hodan2017t}, TUD-L~\cite{hodan2018bop}, IC-BIN~\cite{doumanoglou2016recovering}, ITODD~\cite{drost2017introducing}, HB~\cite{kaskman2019homebreweddb}, YCB-V~\cite{xiang2017posecnn}). 
These datasets comprise 125 unique household and industrial object instances under occlusion, textureless surfaces, illumination changes, and visually similar instances. Following the standard BOP protocol~\cite{sundermeyer2023bop}, we report Average Precision ($AP$) at $IoU \in [0.50:0.05:0.95]$ for detection/segmentation, and Average Recall ($AR$) for downstream 6D pose estimation.

\noindent\textbf{Baselines. \quad} 
We compare \ours with CNOS~\cite{nguyen2023cnos} and SAM-6D~\cite{lin2024sam} front-ends, including their SAM/FastSAM proposal variants when available. MUSE~\cite{cho2026muse} is excluded from the experimental comparison because public code was unavailable at submission.
%
To evaluate whether front-end improvements transfer to pose estimation, we pair each front-end with three downstream pose solvers: GigaPose~\cite{nguyen2024gigapose}, SAM-6D~\cite{lin2024sam}, and FoundationPose~\cite{wen2024foundationpose}. 
%
All baselines are reproduced and evaluated under the same environment, with fixed global hyperparameters and no per-scene/dataset tuning.


\noindent\textbf{Hyperparameters. \quad}
We extract template and proposal features using a frozen DINOv3~\cite{simeoni2025dinov3} ViT-L backbone at a $224 \times 224$ resolution. For each object, we subsample $N'_T = 10$ views at regular angular intervals from the $N_T = 42$ rendered templates and supply them to Qwen-2.5-VL-7B-Instruct~\cite{bai2025qwen25vltechnicalreport} to construct a linguistic memory of $M=4$ short noun phrases. Neither model is fine-tuned.
%
For cue calibration, we compute the appearance entropy using a joint Hue--Saturation histogram with $N_B = 46,080$ bins and set $\tau_{\rm app} = 0.25$. The semantic-priority constant is fixed to $\lambda_{\rm sem} = 3$. 
During inference, we use SAM~3 at a $1008 \times 1008$ resolution using the released checkpoint without further fine tuning. We set the IoU threshold for non maximum suppression of SAM~3 proposals to $\tau_{\rm iou} = 0.95$ and the matching threshold to $\tau_{\rm match}=0.25$, below which proposal object matches are discarded.
All hyperparameters are fixed globally across datasets.

\noindent\textbf{Runtime Implementation.\quad}
All experiments are run on a single NVIDIA RTX PRO 6000 GPU using PyTorch 2.7.0 with CUDA 12.8; SAM 3 and DINOv3 use FP16 and FP32, respectively.
Linguistic memories, cue weights, and template features are computed \textit{once} offline and cached.
During online inference, SAM~3 image features are computed once per scene and shared across queries, while proposal features are extracted in batch. Runtime is measured and averaged over all test images after GPU warm-up under the same hardware setup. 
We will release the \ours implementation upon acceptance. 


\subsection{Main Evaluation on BOP Benchmarks}

\noindent\textbf{2D Object Detection and Segmentation.\quad} 
Table~\ref{tab:detection_segmentation} compares front-end detection and segmentation quality. \ours achieves the highest $AP_{\rm mean}$ on both tasks, reaching 0.553 for detection and 0.543 for segmentation. This improves over the strongest baseline by $+$8.2 pp for detection and $+$6.2 pp for segmentation. The consistent gains in both detection and segmentation indicate that \ours provides more reliable object localization and mask estimation before downstream pose estimation. The contributions of linguistic conditioning, view aggregation, and verification are analyzed separately in the ablation studies in Sec.~\ref{sec:ablation}.
%

\noindent\textbf{6D Pose Estimation Results.\quad} 
Table~\ref{tab:pose_estimation_results} evaluates the effect of front-end quality on three pose solvers. \ours obtains the highest $AR_{\rm mean}$ with all downstream solvers: $0.634$ with GigaPose, $0.726$ with SAM-6D, and $0.748$ with FoundationPose. 
%
Across the 21 solver--dataset combinations, \ours achieves higher AR than the strongest baseline in 17 cases. The exceptions are FoundationPose on T-LESS, TUD-L, ITODD, and HB; the present evaluation does not isolate the causes of these regressions. 
Overall, the results support the utility of the CLON front-end across multiple pose estimators without implying improvement in every setting.

%

\begin{table}[t!]
\centering
\caption{Latency and memory consumption per image across BOP datasets, using SAM-6D as the pose solver. The best and second-best results within each dataset are shown in \textbf{bold} and \underline{underline}, respectively.}
\label{tab:memory_time_eval}
\setlength{\tabcolsep}{4pt}
\resizebox{\columnwidth}{!}{
\begin{tabular}{llccccc}
\toprule
\makecell{Dataset \\ (\# targets)} & Detection & \makecell{\# \\ Proposal}& \makecell{Det \\ Memory (GB)}&  \makecell{Det \\ Time (s)} & \makecell{Pose Est \\ Time (s)} & \makecell{Total \\ Time (s)} \\
\midrule
\multirow{3}{*}{\makecell[l]{LM-O \\ (8 targets)}} 
 & CNOS  & \underline{73} & \textbf{5.20} & 3.31 & \underline{1.07} & \underline{4.38} \\
 & SAM-6D & 112  & 6.55 & \underline{3.26} & 1.75 & 5.01 \\
 & \ours  & \textbf{22} & \underline{5.86} & \textbf{1.02} & \textbf{0.44} & \textbf{1.46} \\
\midrule
\multirow{3}{*}{\makecell[l]{T-LESS \\ (30 targets)}} 
 & CNOS  & \underline{62} & \textbf{5.20}& \textbf{2.98} & \underline{0.80} & \underline{3.78} \\
 & SAM-6D & 79 & 6.74 & 4.08 & 1.18 & 5.27 \\
 & \ours  & \textbf{12} & \underline{6.37} & \underline{3.25} & \textbf{0.24} & \textbf{3.49} \\
\midrule
\multirow{3}{*}{\makecell[l]{TUD-L \\ (3 targets)}} 
 & CNOS  & \underline{35} & \underline{5.20} & \underline{2.13} & \underline{0.39} & \underline{2.53} \\
 & SAM-6D & 62 & 5.32 & 2.32 & 0.76 & 3.08 \\
 & \ours  & \textbf{5} & \textbf{5.02} & \textbf{0.46} & \textbf{0.14} & \textbf{0.60} \\
\midrule
\multirow{3}{*}{\makecell[l]{IC-BIN \\ (2 targets)}} 
 & CNOS  & \underline{45} & \textbf{5.20} & \underline{2.67} & \underline{0.62} & \underline{3.30} \\
 & SAM-6D &  63 & 5.28 & 2.97 & 0.97 & 3.94 \\
 & \ours  & \textbf{15} & 5.61 & \textbf{0.33} & \textbf{0.31} & \textbf{0.64} \\
\midrule
\multirow{3}{*}{\makecell[l]{ITODD \\ (28 targets)}} 
 & CNOS  & \textbf{18} & \textbf{5.21} & \textbf{1.99} & \textbf{0.30} & \textbf{2.29} \\
 & SAM-6D & 27  & 6.82 & 3.11 & 0.44 & 3.55 \\
 & \ours  & \underline{20} & \underline{6.15} & \underline{3.09} & \underline{0.39} & \underline{3.48} \\
\midrule
\multirow{3}{*}{\makecell[l]{HB \\ (33 targets)}} 
 & CNOS  & \underline{46} & \textbf{5.20} & \textbf{2.43} & \underline{0.72} & \textbf{3.15} \\
 & SAM-6D & 70 & 6.53 & \underline{3.20} & 1.08 & 4.29 \\
 & \ours  & \textbf{13} & \underline{6.45} & 3.34 & \textbf{0.62} & \underline{3.96} \\
\midrule
\multirow{3}{*}{\makecell[l]{YCBV \\ (21 targets)}} 
 & CNOS  & \underline{37} & \textbf{5.20}& \underline{2.49} & \underline{0.58} & \underline{3.07} \\
 & SAM-6D & 58 & 6.06 & 2.87 & 0.89 & 3.76 \\
 & \ours & \textbf{8}& \underline{5.79} & \textbf{2.15} & \textbf{0.28} & \textbf{2.43} \\
\bottomrule
\end{tabular}
}
\end{table}

\begin{table}[t]
\centering
\caption{Proposal count and class-agnostic mask recall after proposal consolidation and before proposal--object verification. Proposal count is reported for all seven datasets, while $AR_{\rm mean}$ is reported only for the five datasets with available ground-truth annotations and averages recall over mask IoU thresholds $[0.50:0.05:0.95]$ using all retained proposals and all ground-truth instances. Dashes denote unavailable recall measurements for ITODD and HB.}
\label{tab:proposal_recall}
\resizebox{\columnwidth}{!}{%
\begin{tabular}{ll|ccccccc}
\toprule
\textbf{Method} & \textbf{Metric} & \textbf{LM-O} & \textbf{T-LESS} & \textbf{TUD-L} & \textbf{IC-BIN} & \textbf{ITODD} & \textbf{HB} & \textbf{YCB-V} \\
\midrule

SAM-6D
& Count
& 174 & 107 & 95 & 108 & 36 & 89 & 80 \\
(SAM) 
& $AR_{\rm mean}$
& 0.650 & 0.769 & 0.763 & 0.495 & -- & -- & 0.772 \\
\midrule

\ours
& Count
& 23 & 14 & 5 & 16 & 21 & 14 & 9 \\
(SAM 3 + LM)
& $AR_{\rm mean}$
& 0.654 & 0.818 & 0.778 & 0.511 & -- & -- & 0.837 \\
\bottomrule
\end{tabular}%
}
\end{table}

\noindent\textbf{Computational Efficiency.\quad}
Table~\ref{tab:memory_time_eval} reports the per-image latency and memory consumption across all seven BOP datasets, with SAM-6D~\cite{lin2024sam} as the downstream pose solver for all methods. The reported end-to-end latency includes both front-end detection and downstream 6D pose estimation. We measure CNOS and SAM-6D with the SAM proposal generator and use the same SAM-6D pose solver for all front-ends. SAM-6D (SAM) is the strongest complete mean baseline in pose evaluations (Table~\ref{tab:pose_estimation_results}).  

Although the detection time of \ours tends to increase with the number of target objects, \ours substantially reduces the number of proposals forwarded to downstream pose estimation, from $67.3$ proposals per image with SAM-6D to $13.6$ on average. 
Averaged across the seven datasets, this reduces the pose estimation time from $1.01$ s to $0.35$ s and the total latency from $4.13$ s to $2.29$ s. \ours achieves lower total latency than the measured SAM-6D configuration on all seven datasets, while remaining comparable to CNOS, with lower latency on five of the seven datasets. Overall, these results show that the more compact detection output of \ours reduces downstream pose computation while maintaining a favorable tradeoff in accuracy and efficiency.

For offline onboarding, \ours takes $73.7$\,s per object set on average across datasets, compared with $34.9$\,s for SAM-6D. 
Timing excludes CAD rendering and includes rendered-image loading and template feature extraction for both methods, with linguistic memory generation and cue weight computation additionally included for \ours.
Qwen2.5-VL-7B-Instruct~\cite{bai2025qwen25vltechnicalreport} reaches $21$\,GB peak memory during onboarding and is not used during online inference. The resulting memories, cue weights, and template features are cached for reuse, while Table~\ref{tab:memory_time_eval} reports online costs separately.

\noindent\textbf{Segmentation Proposal Coverage.\quad}
Table~\ref{tab:proposal_recall} evaluates mask proposals given by the proposal generator (SAM 3 + LM) before proposal–object verification on the five datasets for which ground-truth annotations are available for recall evaluation. Across the five datasets, the equal-dataset mean proposal count decreases from 112.8 to 13.4, an 88.1\% reduction, while class-agnostic mask recall increases from 0.690 to 0.720. Thus, compactness is already achieved after proposal consolidation, before subsequent verification and filtering.
This comparison evaluates the complete proposal paths, while Table~\ref{tab:component_ablation} and~\ref{tab:prompt_ablation}(A) separately examine linguistic conditioning with the verifier fixed.


\subsection{Ablation Study}
\label{sec:ablation}

\begin{table}[t]
\centering
\caption{Ablation of proposal generation, linguistic memory, view aggregation, and verification. (A) fixes the descriptor, \ours verifier, cue weights, and postprocessing while varying the proposal generator and prompt construction. (B) fixes SAM~3 and LM and compares semantic-first verification (SAM-6D), an all-object reference (SAM-6D-All) and \ours. SAM-6D-All and \ours share the proposal–object cue scores, assignment rule, and postprocessing; only their cue weights differ. G-DINO denotes Grounding DINO followed by SAM.}
\label{tab:component_ablation}
\small
\setlength{\tabcolsep}{6pt}

\resizebox{\columnwidth}{!}{%
\begin{tabular}{lcc|cc}
\toprule

\multicolumn{3}{c|}{\makecell{\textbf{ (A) Proposal Generation \& Prompt}\\ \scriptsize (\ours verifier fixed)}}
&
\multicolumn{2}{c}{\makecell{\textbf{ (B) Verification Policy}\\ \scriptsize (SAM~3 + LM + three cues fixed)}} \\

\cmidrule(lr){1-3}
\cmidrule(lr){4-5}

\textbf{Proposal Generator} & \textbf{Prompt} & $AP_{\text{mean}}$
& \textbf{Verification} & $AP_{\text{mean}}$ \\

\midrule

SAM & -- & 0.440
& SAM-6D & 0.540 \\

G-DINO & Object & 0.312
& SAM-6D-All & 0.535 \\

G-DINO & Multi-view LM & 0.449
& \textbf{\ours} & \textbf{0.553} \\

SAM~3 & Object & 0.412
& -- & -- \\

SAM~3 & Single-view LM & 0.536
& -- & -- \\

\textbf{SAM~3} & \textbf{Multi-view LM} & \textbf{0.553}
& -- & -- \\

\bottomrule
\end{tabular}%
}
\end{table}

\begin{table}[t]
\centering
\caption{Ablation study of linguistic prompt composition and prompt modality for \ours. The left block evaluates different compositions of linguistic prompts, while the right block compares linguistic and image-based prompt modalities.
For each ablation, all other components (SAM~3 and Cue Weight) are fixed to isolate the effect of the prompt. Here, Object denotes the generic prompt set $T_{\rm gen}$.}
\label{tab:prompt_ablation}
\small
\setlength{\tabcolsep}{3pt}
\begin{tabular}{lc|lc}
\toprule
\textbf{(A) Linguistic Prompt} & $AP_{\text{mean}}$ & \textbf{ (B) Prompt Modality} & $AP_{\text{mean}}$ \\
\midrule
Object & 0.412 & Image & 0.008 \\
Object + Color & 0.488 & Image + LM & \textbf{0.553} \\
Object + Top-1 & 0.525 & \textbf{LM (\ours)} & \textbf{0.553} \\
Object + Top-1 + Color & 0.544 & -- & -- \\
\textbf{LM (\ours)} & \textbf{0.553} & -- & -- \\
\bottomrule
\end{tabular}
\end{table}

\noindent\textbf{LM and View Aggregation.} \quad 
Table~\ref{tab:component_ablation}(A) separates proposal generation and linguistic conditioning from verification. With the CLON verifier fixed, generic SAM~3 achieves 0.412 $AP_{\rm mean}$, compared with 0.440 for SAM proposals, showing that replacing the proposal generator alone does not explain the gain of \ours. Adding LM improves Grounding DINO + SAM from 0.312 to 0.449 and SAM~3 from 0.412 to 0.553, demonstrating that the benefit of linguistic conditioning is not limited to SAM~3. 
We further compare frontal single-view and multi-view ($N'_T=10$) LM while keeping the number of generated phrases fixed.
Multi-view onboarding improves $AP_{\rm mean}$ from 0.536 to 0.553, including gains from 0.300 to 0.358 on IC-BIN and from 0.444 to 0.465 on ITODD.
This improvement likely arises because multiple viewpoints allow concise noun phrases to capture more accurate and complementary object information. 


\noindent\textbf{Verification Policy and Cue Weighting.}\quad
Table~\ref{tab:component_ablation}(B) fixes SAM~3 and LM and separates the effect of object assignment from cue weighting.
In the original SAM-6D policy, each proposal is first assigned to the object with the highest semantic score, after which the selected proposal and object pair is scored as
\[S_{\rm SAM6D}=\frac{S_{\rm sem}+S_{\rm app}+r_{\rm vis} S_{\rm geo}}{2 + r_{\rm vis}},\] 
where $r_{\rm vis}$ denotes the estimated visible ratio. SAM-6D-All removes the semantic first assignment and applies the same SAM-6D fusion rule to every proposal and object pair before selecting the object with the highest fused score, but this modification slightly lowers performance to 0.535 $AP_{\rm mean}$. With the cue scores, all object assignment rule, thresholds, and postprocessing fixed to SAM-6D-All, replacing the SAM-6D fusion rule with the onboarding derived cue weights of \ours raises $AP_{\rm mean}$ to 0.553 (+1.8 pp). 
The resulting improvement shows that onboarding derived cue weights improve verification of the fixed proposal set.

\begin{figure}[t!]
  \centering
  \includegraphics[width=\linewidth]{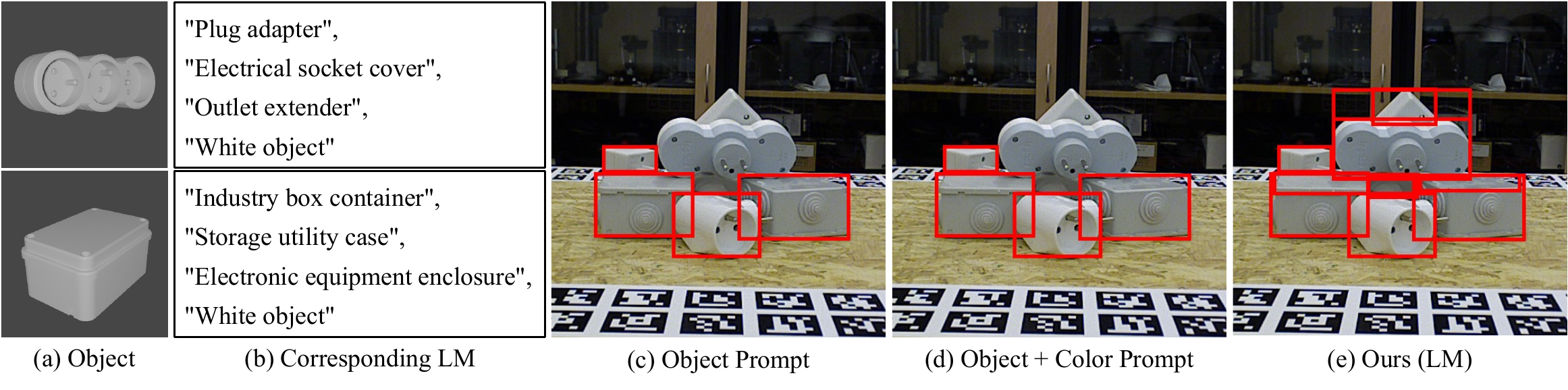}
  \caption{Qualitative results of LM on T-LESS. 
  (a) The template image of target object (b) The LM generated for each object (c)-(e) SAM 3 segmentation with the generic prompt set, generic + color prompts, and our full LM, respectively.
  }
  \label{fig:qualitative_lm}
\end{figure}

\noindent\textbf{Prompt Composition and Modality.\quad}
Table~\ref{tab:prompt_ablation}(A) analyzes how the linguistic information should be constructed for proposal generation.
Here, \textit{Object} denotes the generic prompt set $T_{\rm gen}$. \textit{Color} provides the dominant color of each target object, while \textit{Top-1} uses only the first noun phrase generated by Qwen for each object. \textit{LM} uses the full prompt set $Q=T_{\rm gen}\cup\bigcup_{c\in\mathcal{C}}L_c$, combining the generic prompts with all generated object-specific phrases.
The generic prompt set alone performs poorly, while adding object-specific information progressively improves performance. The strongest restricted prompt, Top-1 + Color, reaches 0.544 $AP_\text{mean}$, and the full LM further improves performance to 0.553. This indicates that richer object descriptions provide additional guidance beyond generic or restricted linguistic prompts.
%
This effect is also evident on texture-less objects. Fig.~\ref{fig:qualitative_lm} shows that LM can recover heavily occluded T-LESS targets missed by generic or color-augmented prompts.


Table~\ref{tab:prompt_ablation}(B) evaluates a single-view composite-image exemplar: a front-view rendering is inserted into the test image, and its bounding box is supplied to SAM 3. This construction achieves 0.008 AP and provides no additional gain when combined with LM. This result applies only to the evaluated construction, not to visual exemplars generally. Under the evaluated configuration, LM provides stronger guidance for target proposal generation.

\begin{figure}[t]
    \centering

    \begin{minipage}{0.57\linewidth}
        \centering
        \includegraphics[width=\linewidth]{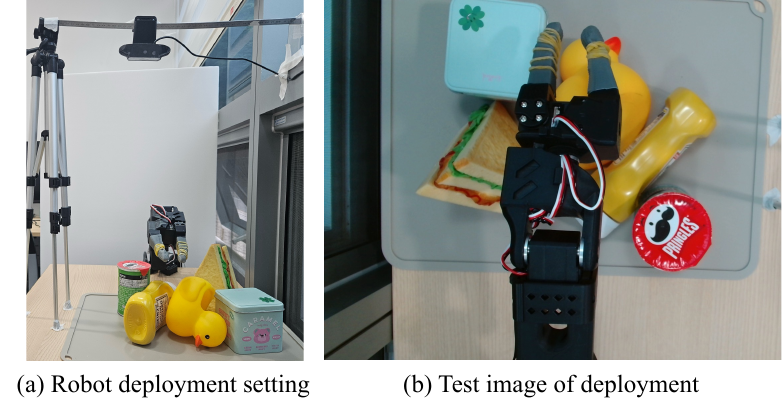}
    \end{minipage}
    \hfill
    \begin{minipage}{0.4\linewidth}
        \centering
        \scriptsize
        \setlength{\tabcolsep}{2pt}
        \resizebox{\linewidth}{!}{%
        \begin{tabular}{lcc}
            \toprule
            Method & $AP_{\text{mean}}$ & Grasp (\%) \\
            \midrule
            CNOS~\cite{nguyen2023cnos} & 0.514 & 12.0 \\
            SAM-6D~\cite{lin2024sam} & 0.284 & 18.0 \\
            \midrule
            \textbf{\ours} & \textbf{0.811} & \textbf{36.0} \\
            \bottomrule
        \end{tabular}%
        }
        
        \vspace{1mm}
        \tiny (c) Quantitative results
    \end{minipage}
    \caption{Real-world robot deployment under clutter and occlusion. 
    (a) Experimental setup, (b) example test image, and 
    (c) quantitative object detection and grasping success rate.}
    \label{fig:robot_deployment}
\end{figure}

\begin{figure}[t]
    \centering
    \includegraphics[width=\linewidth]{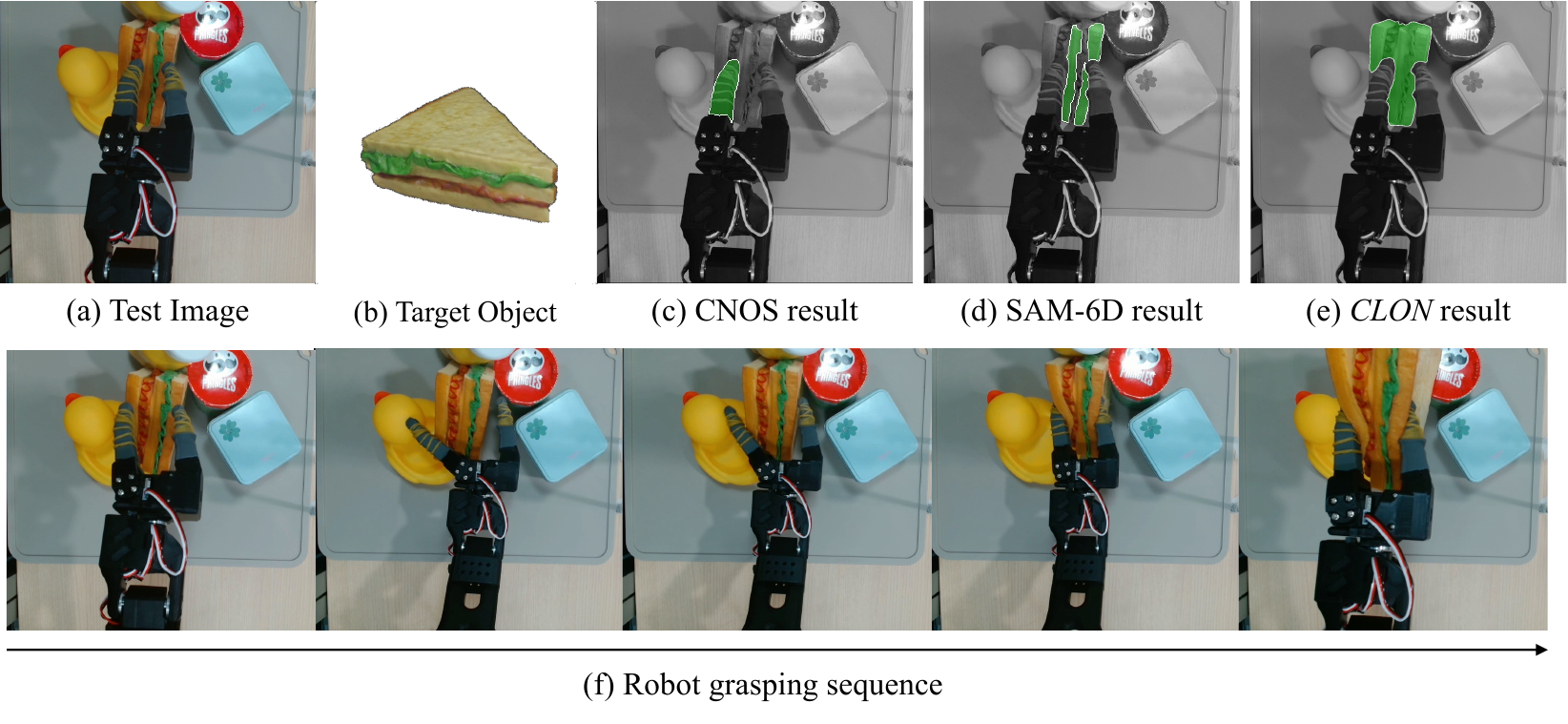}
    \caption{Qualitative results of real-world robot deployment. (a) Test image, (b) target object, (c–e) detection results of CNOS, SAM-6D, and \ours, and (f) robot grasping sequence using the \ours detection result.}
    \label{fig:robot_grasp_result}
\end{figure}

\subsection{Real-World Robot Evaluation}

We further evaluate whether improved front-end localization transfers to real-world grasping performance (i.e., successfully grasping and lifting a target object to the top position). Our setup uses an SO-101 robot arm with a standard gripper and a Logitech C920E webcam mounted above workspace to capture fixed bird's-eye-view RGB images (Figs.~\ref{fig:robot_deployment}(a) and (b)). We evaluate five household objects from SenseShift6D~\cite{han2026senseshift6d}--sandwich, tin case, Pringles can, duck, and mustard bottle--in cluttered scenes with occlusions caused by surrounding objects and the robot arm. For each object, we conduct 20 trials, resulting in 100 test scenes in total for each method, and manually annotate the ground-truth bounding boxes for detection evaluation. The captured RGB image is transmitted to a remote GPU server, where each method localizes the target object. The center of the predicted bounding box is then converted to the robot workspace and passed to an inverse-kinematics (IK) controller to execute an open-loop grasping trajectory.

For SAM-6D, which requires depth, we use Depth Anything 3~\cite{depthanything3} to estimate depth from the same RGB input. For \ours, the onboarding cue weights satisfy $e_{\rm app}>\tau_{\rm app}$ for this object set, resulting in $(w_{\rm sem},w_{\rm app},w_{\rm geo})=(1,0,0)$. Accordingly, this experiment compares the deployed pipelines as configured, rather than isolating linguistic conditioning under identical verification inputs.
This experiment evaluates the transfer from target localization to grasp execution and does not use downstream 6D pose estimation.

As shown in Fig~\ref{fig:robot_deployment}(c), \ours achieves the highest detection performance with an $AP_{\text{mean}}$ of 0.811, and increases grasp success from 18.0\% for the strongest baseline to 36.0\%. 
Fig.~\ref{fig:robot_grasp_result} illustrates a representative case in which CNOS and SAM-6D produce incorrect or fragmented detections under occlusion, whereas \ours correctly localizes the target object and enables a successful grasp and lift. These results provide evidence that improved localization can transfer to higher grasp success in the tested setup. However, the study does not evaluate 6D-pose-based control or separate the remaining failures into localization and execution errors. 

\section{Conclusion}
\label{sec:conclusion}

We presented \ours, a training-free front-end that converts multi-view CAD evidence into reusable linguistic conditioning for proposal generation. Its central intervention is to use onboarding before candidate matching, while template verification assigns identity and template-derived cue weights provide a complementary improvement. Across seven BOP datasets and three pose solvers, the experiments support improved localization and mean pose accuracy, together with a compact pre-verification candidate set and lower measured average latency than SAM-6D (SAM). A real-world robot study further shows that these localization improvements transfer to higher grasp success under clutter and occlusion.

\noindent\textbf{Limitations.\quad}
Linguistic descriptions may not capture distinctions between visually similar or uncommon objects. Cue weights are fixed during onboarding and do not adapt to scene-dependent reliability. The robot study evaluates 2D localization-to-grasp transfer rather than pose-based control. 
Future work will investigate more discriminative onboarding representations and scene-adaptive verification.




\bibliographystyle{IEEEtran}
\bibliography{main}  

\end{document}